\documentclass[11pt,letterpaper]{article}
\usepackage[T1]{fontenc}
\usepackage[utf8]{inputenc}
\usepackage{authblk}
\usepackage{ijcnlp2017}
\usepackage{times}
\usepackage{latexsym}

\usepackage{amsmath,scalerel}

\usepackage{amsfonts}
\usepackage{multirow}
\usepackage{graphicx}
\usepackage{enumerate}
\usepackage{tikz}
\usetikzlibrary{bayesnet}
\usepackage{soul}
\usepackage{linguex}
\usepackage{esvect}
\usepackage{url}
\usepackage{rotating}
\usepackage{bm}
\usepackage{amssymb}
\usepackage{booktabs}
\usepackage{float}
\usepackage{rst}
\usepackage{subfigure}
\usepackage{amssymb}
\usepackage{setspace}
\ijcnlpfinalcopy

\def\ijcnlppaperid{***}

\title{Leveraging Discourse Information Effectively for Authorship Attribution\thanks{The first two authors make equal contribution.}}
\author{\textbf{Su Wang}$^\spadesuit$\quad\textbf{Elisa Ferracane}$^\spadesuit$\quad\textbf{Raymond J. Mooney}$^\clubsuit$ \\
 {$^\spadesuit$Department of Linguistics, $^\clubsuit$Department of Computer Science} \\
 {The University of Texas at Austin} \\
 {\tt shrekwang@utexas.edu, elisa@ferracane.com mooney@cs.utexas.edu}
 } 

\begin{document}

\maketitle

\begin{abstract}
We explore techniques to maximize the effectiveness of discourse information in the task of authorship attribution. 
We present a novel method to embed discourse features in a Convolutional Neural Network text classifier, which achieves a state-of-the-art result by a substantial margin.
We empirically investigate several featurization methods to understand the conditions under which discourse features contribute non-trivial performance gains, and analyze discourse embeddings. 
\end{abstract}

\vspace{-.5em}
\begin{table*}[t]
\centering
\begin{tabular}{p{15.5cm}}
\hline(1) [My father]$_S$ was a clergyman of the north of England, [who]$_O$ was deservedly respected by all who knew [him]$_O$; and, in his younger days, lived pretty comfortably on the joint income of a small incumbency and a snug little property of his own. \\
(2) [My mother]$_S$, who married [him]$_O$ against the wishes of her friends, was a squire's daughter, and a woman of spirit. \\  
(3) In vain it was represented to [her]$_X$, that if [she]$_S$ became [the poor parson's]$_X$ wife, [she]$_S$ must relinquish her carriage and her lady's-maid, and all the luxuries and elegancies of affluence; which to [her]$_X$ were little less than the necessaries of life.\\\hline
\end{tabular}
\vspace{-.8em}
\caption{\label{tab:sents} Excerpt of 19$^{th}$-century novel where sentences are labeled with the salient entities and their grammatical relations (subject \textbf{s}, object \textbf{o}, other relation \textbf{x}). A salient entity is a noun phrase coreferred to at least two times in a document.}
\end{table*}

\begin{table*}[t]
\centering
\begin{tabular}{lllllllllllllllll}\hline
  & ss &so &sx &s- &os &oo &ox &o- &xs &xo &xx &x- &-s &-o &-x &--\\\hline 
d$_1$& 0.25 &0.25 &0 &0 &0 &0 &0.25 &0 &0 &0 &0 &0 &0.25 &0 &0 &0\\\hline 
\end{tabular}
\vspace{-.5em}
\caption{\label{tab:vector} The probability vector for the excerpt in Table \ref{tab:sents} capturing transition probabilities of length 2.}
\end{table*}

\section{Introduction}
\label{sec:introduction}
Authorship attribution (AA) is the task of identifying the author of a text, given a set of author-labeled training texts. This task typically makes use of stylometric cues at the surface lexical and syntactic level \cite{Stamatatos:15}, although \citet{Feng:14} and \citet{feng2015} go beyond the sentence level, showing that discourse information can help. However, they achieve limited performance gains and lack an in-depth analysis of discourse featurization techniques. More recently, convolutional neural networks (CNNs) have demonstrated considerable success on AA relying  only on character-level $n$-grams \cite{Ruder:16,shrestha2017}. The strength of these models is evidenced by findings that traditional stylometric features such as word $n$-grams and POS-tags do not improve, and can sometimes even hurt performance \citep{Ruder:16,Sari:17}. However, none of these CNN models make use of discourse.

Our work builds upon these prior studies by exploring an effective method to (i) featurize the discourse information, and (ii) integrate discourse features into the best text classifier (i.e., CNN-based models), in the expectation of achieving state-of-the-art results in AA.


\citet{Feng:14} (henceforth F\&H14) made the first comprehensive attempt at using discourse information for AA. They employ an entity-grid model, an approach introduced by \citet{Barzilay:08} for the task of ordering sentences. This model tracks how the grammatical relations of salient entities (e.g., \texttt{subj}, \texttt{obj}, etc.) change between pairs of sentences in a document, thus capturing a form of discourse coherence. The grid is summarized into a vector of transition probabilities. However, because the model only records the transition between two consecutive sentences at a time, the coherence is \textit{local}. \citet{feng2015} (henceforth F15) further extends the entity-grid model by replacing grammatical relations with discourse relations from Rhetorical Structure Theory \citep[RST]{Mann:88}. Their study uses a linear-kernel SVM to perform pairwise author classifications, where a non-discourse model captures lexical and syntactic features. They find that adding the entity-grid with grammatical relations enhances the non-discourse model by almost 1\% in accuracy, and using RST relations provides an improvement of 3\%. The study, however, works with only one small dataset and their models produce overall unremarkable performance ($\sim$85\%). 
\citet{Ji:17} propose an advanced Recursive Neural Network (RecNN) architecture to work with RST in the more general area of text categorization and present impressive results.
However, we suspect that the massive number of parameters of RecNNs would likely cause overfitting when working with smaller datasets, as is often the case in AA tasks. 

In our paper, we opt for a state-of-the-art character bigram CNN classifier \cite{shrestha2017}, and investigate various ways in which the discourse information can be featurized and integrated into the CNN. 
Specifically,
\begin{itemize}
\item \emph{Featurization}. We attempt to capture a more \emph{global} discourse coherence by modeling the entire sequence of relations in a document for every salient entity, instead of only the relations between pairs of sentences. 
\item \emph{Feature integration}. Using a neural network architecture allows us to explore embedding the relations from the entity-grid model\footnote{\newcite{Nguyen:17} are the first to propose applying embeddings in modeling local coherence (for the coherence judgment task). Our methods roughly subsume theirs, which correspond to our GR CNN2-DE (global) model (Section \ref{sec:models}). This scheme did not come out on top in our AA tasks.}, rather than only exploiting a vector of relation probabilities. 
\end{itemize}


We explore these questions using two approaches to represent salient entities: grammatical relations, and RST discourse relations. We apply these models to datasets of varying sizes and genres, and find that adding any discourse information improves AA consistently on longer documents, but has mixed results on shorter documents. Further, embedding the discourse features in a parallel CNN at the input end yields better performance than concatenating them to the output layer as a feature vector (Section \ref{sec:models}). The global featurization is more effective than the local one.
We also show that SVMs, which can only use discourse probability vectors, neither produce a competitive performance (even with fine-tuning), nor generalize in using the discourse information effectively.

\section{Background}
\label{sec:background}
\textbf{Entity-grid model.} Typical lexical features for AA are relatively superficial and restricted to within the same sentence. F\&H14 hypothesize that discourse features beyond the sentence level also help authorship attribution. In particular, they propose an author has a particular style for representing entities across a discourse. Their work is based on the entity-grid model of \citet{Barzilay:08} (henceforth B\&L). 

The entity-grid model tracks the grammatical relation (\texttt{subj}, \texttt{obj}, etc.) that salient entities take on throughout a document as a way to capture local coherence . A salient entity is defined as a noun phrase that co-occurs at least twice in a document. Extensive literature has shown that subject and object relations are a strong signal for salience and it follows from the Centering Theory that you want to avoid rough shifts in the center \cite{grosz1995,strube1999}. B\&L thus focus on whether a salient entity is a subject (\textbf{s}), object (\textbf{o}), other (\textbf{x}), or is not present (\textbf{-}) in a given sentence, as illustrated in Table \ref{tab:sents}. Every sentence in a document is encoded with the grammatical relation of all the salient entities, resulting in a grid similar to Table \ref{tab:grid}. 

\begin{table}[h]
\centering
\begin{tabular}{lll}
       & \begin{turn}{45}father\end{turn}  
       & \begin{turn}{45}mother\end{turn}  \\ \cline{2-3} 
\multicolumn{1}{l|}{(1)} & \multicolumn{1}{l|}{s} & \multicolumn{1}{l|}{-} \\ \cline{2-3} 
\multicolumn{1}{l|}{(2)} & \multicolumn{1}{l|}{o} & \multicolumn{1}{l|}{s} \\ \cline{2-3} 
\multicolumn{1}{l|}{(3)} & \multicolumn{1}{l|}{x} & \multicolumn{1}{l|}{s} \\ \cline{2-3} 
\end{tabular}
\vspace{-.5em}
\caption{\label{tab:grid} The entity grid for the excerpt in Table \ref{tab:sents}, where columns are salient entities and rows are sentences. Each cell contains the grammatical relation of the given entity for the given sentence (subject \textbf{s}, object \textbf{o}, another grammatical relation \textbf{x}, or not present \textbf{-}). If an entity occurs multiple times in a sentence, only the highest-ranking relation is recorded.}
\end{table}

The local coherence of a document is then defined on the basis of local entity transitions. A local entity transition is the sequence of grammatical relations that an entity can assume across $n$ consecutive sentences, resulting in \{s,o,x,-\}$^n$ possible transitions. Following B\&L, F\&H14 consider sequences of length $n$=$2$, that is, transitions between two consecutive sentences, resulting in $4^2$=$16$ possible transitions. The probability for each transition is then calculated as the frequency of the transition divided by the total number of transitions. This step results in a single probability vector for every document, as illustrated in Table \ref{tab:vector}.

B\&L apply this model to a sentence ordering task, where the more coherent option, as evidenced by its transition probabilities, was chosen. In authorship attribution, texts are however assumed to already be coherent. F\&H14 instead hypothesize that an author unconsciously employs the same methods for describing entities as the discourse unfolds, resulting in discernible transition probability patterns across multiple of their texts. Indeed, F\&H14 find that adding the B\&L vectors increases the accuracy of AA by almost 1\% over a baseline lexico-syntactic model.
\medskip

\begin{figure}
\vspace{-5em}
\includegraphics[width=\linewidth]{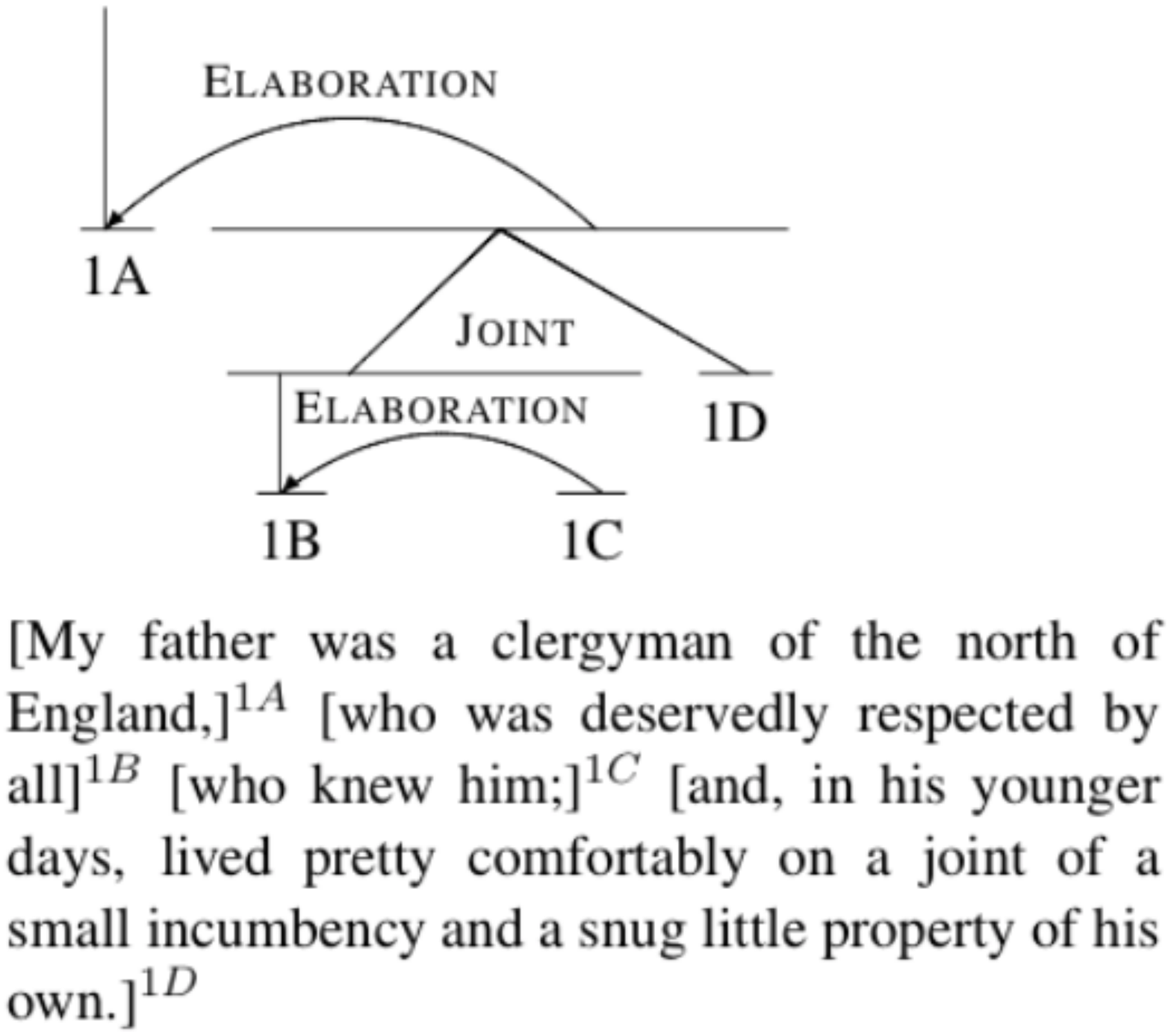}
\vspace{-7.5em}
\caption{RST tree for the first sentence of the excerpt in Table \ref{tab:sents}.}
\label{fig:rstTree}
\end{figure}

\noindent\textbf{RST discourse relations.} F15 extends the notion of tracking salient entities to RST. Instead of using grammatical relations in the grid, RST discourse relations are specified. An RST discourse relation defines the relationship between two or more elementary discourse units (EDUs), which are spans of text that typically correspond to syntactic clauses. In a relation, an EDU can function as a nucleus (e.g., \texttt{result.N}) or as a satellite (e.g., \texttt{summary.S}). All the relations in a document then form a tree as in Figure \ref{fig:rstTree}.\footnote{For reasons of space, only the first sentence of the excerpt is illustrated.}

%

F15 finds that RST relations are more effective for AA than grammatical relations. In our paper, we populate the entity-grid in the same way as F15's ``Shallow RST-style'' encoding, but use fine-grained instead of coarse-grained RST relations, and do not distinguish between intra-sentential and multi-sentential RST relations, or salient and non-salient entities. 
We explore various featurization techniques using the coding scheme. 
\medskip

\noindent\textbf{CNN model.}
\newcite{shrestha2017} propose a convolutional neural network formulation for AA tasks (detailed in Section \ref{sec:models}).
They report state-of-the-art performance on a corpus of Twitter data \citep{Schwartz:13}, and compare their models with alternative architectures proposed in the literature:
(i) SCH: an SVM that also uses character n-grams, among other stylometric features \citep{Schwartz:13};
(ii) LSTM-2: an LSTM trained on bigrams \citep{Tai:15};
(iii) CHAR: a \emph{Logistic Regression} model that takes character n-grams \citep{Stamatatos:09};
(iv) CNN-W: a CNN trained on word embeddings \citep{Kalchbrenner:14}.
The authors show that the model CNN2\footnote{\citet{shrestha2017} test two variants of CNN models: \textbf{CNN1}/\textbf{CNN2} for unigram/bigram character CNNs respectively.} produces the best performance overall.
\newcite{Ruder:16} apply character $n$-gram CNNs to a wide range of datasets, providing strong empirical evidence that the architecture generalizes well. 
Further, they find that including word $n$-grams in addition to character $n$-grams reduces performance, which is in agreement with \citet{Sari:17}'s findings. 

\begin{figure*}
\includegraphics[width=\linewidth]{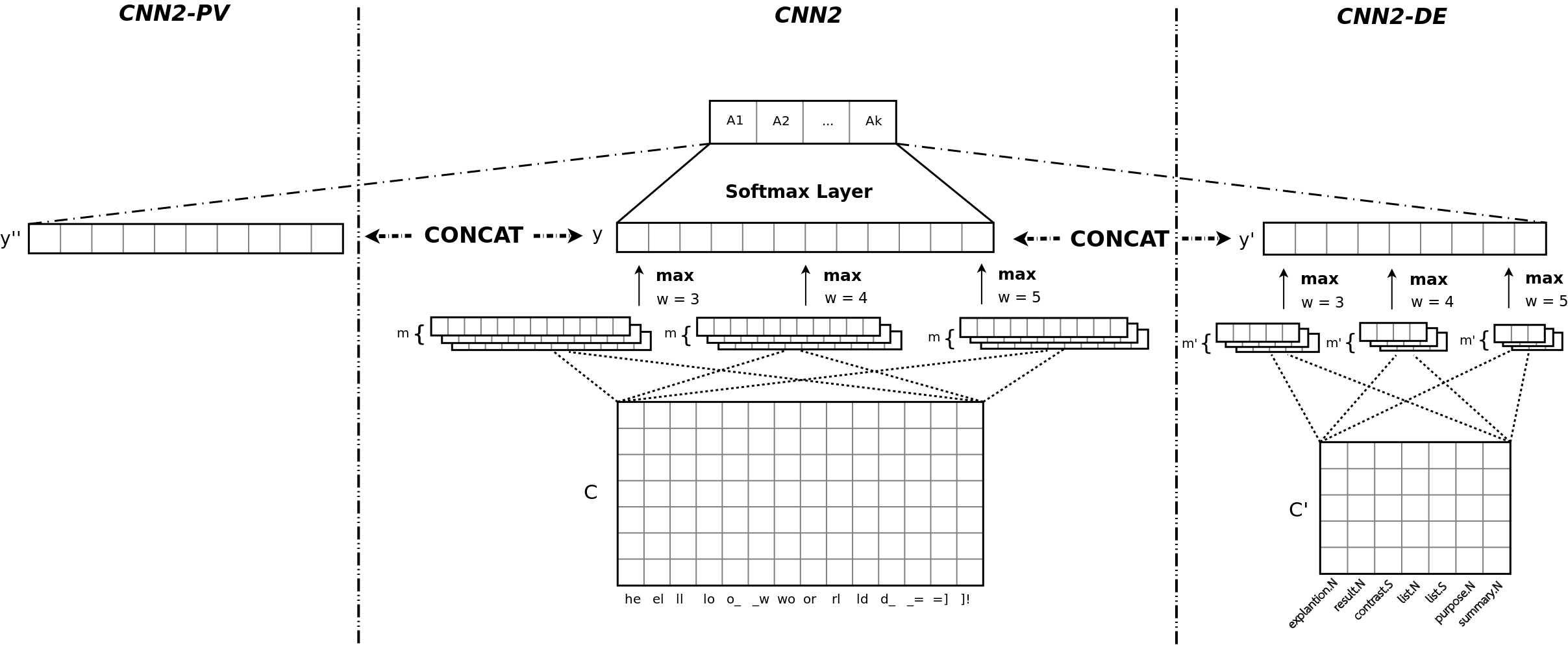}
\vspace{-1.5em}
\caption{The bigram character CNN models}
\label{fig:cnn2}
\end{figure*}

\section{Models}
\label{sec:models}


Building on \newcite{shrestha2017}'s work,
 we employ their character-bigram CNN (\textbf{CNN2})\footnote{Our preliminary experiments found that using character $n$-gram orders higher than 2 performed worse, likely due to the increased number of features and overfitting.}, and propose two extensions which utilize discourse information:
(i) CNN2 enhanced with relation \emph{probability vectors} (\textbf{CNN2-PV}), and (ii) CNN2 enhanced with \emph{discourse embeddings} (\textbf{CNN2-DE}). 
The CNN2-PV allows us to conduct a comparison with F\&H14 and F15, which also use relation probability vectors. 
\medskip

\noindent
\textbf{CNN2.}
CNN2 is the baseline model with no discourse features. Illustrated in Figure \ref{fig:cnn2} (center), it consists of (i) an embedding layer, (ii) a convolution layer, (iii) a max-pooling layer, and (iv) a softmax layer.
We briefly sketch the processing procedure and refer the reader to \citep[Section 2]{shrestha2017} for mathematical details.

The network takes a sequence of character bigrams $\bm{x} = \langle x_1,\ldots,x_l\rangle$ as input, and outputs a multinomial $\phi$ over class labels as the prediction.
The model first looks up the embedding matrix to produce a sequence of embeddings for $\bm{x}$ (i.e., the matrix $C$), then pushes the embedding sequence through convolutional filters of three bigram-window sizes $w=3,4,5$, each yielding $m$ feature maps. 
We then apply the \emph{max-over-time} pooling \cite{Collobert:11} to the feature maps from each filter, and concatenate the resulting vectors to obtain a single vector $\bm{y}$, which then goes through the softmax layer to produce predictions.

\medskip


\noindent
\textbf{CNN2-PV}.
This model (Figure \ref{fig:cnn2}, left+center) featurizes discourse information into a vector of relation probabilities. 
In order to derive the discourse features, an entity grid is constructed by feeding the document through an NLP pipeline\footnote{Using neural coreference resolver, dependency parser in Stanford Core NLP \cite{Clark:2016}.} to identify salient entities. Two flavors of discourse features are created by populating the entity grid with either 
(i) grammatical relations (GR) or (ii) RST discourse relations\footnote{Using RST Parser from \citet{Ji:14}.} (RST). 
The GR features are represented as \emph{grammatical relation transitions} derived from the entity grid,
e.g., $\langle\texttt{sx,xs,ss,\ldots}\rangle$.
The RST features are represented as \emph{RST discourse relations} with their nuclearity,
e.g., $\langle\texttt{definition.N,attribution.S,\ldots}\rangle$.
The probability vectors are then distributions over relation types. For GR, the vector is a distribution over all the entity role transitions, i.e., $\langle p(\texttt{sx}),p(\texttt{xs}),p(\texttt{ss}),\ldots\rangle$ (see Table \ref{tab:vector}). For RST, the vector is a distribution over all the RST discourse relations, i.e., $\langle p(\texttt{definition.N}),p(\texttt{attribution.S}),\ldots\rangle$
Denoting a feature as such with $\bm{y}''$, we construct the pooling vector $\bm{y}$ for the char-bigrams, and concatenate $\bm{y}''$ to $\bm{y}$ before feeding the resulting vector to the softmax layer. 
\medskip

\noindent
\textbf{CNN2-DE}.
In this model (Figure \ref{fig:cnn2}, center+right), we embed discourse features in high-dimensional space (similar to char-bigram embeddings).
Let $\bm{z} = \langle z_1,\ldots,z_{l'}\rangle$ be a sequence of discourse features\footnote{The sequence comes in two variants, depending on the featurization technique, see Section 4.2.}, we treat it in a similar fashion to the char-bigram sequence $\bm{x}$, i.e. feeding it through a ``parallel'' convolutional net (Figure \ref{fig:cnn2} right). 
The operation results in a pooling vector $\bm{y}'$.
We concatenate $\bm{y}'$ to the pooling vector $\bm{y}$ (which is constructed from $\bm{x}$) then feed $[\bm{y};\bm{y}']$ to the softmax layer for the final prediction.


\medskip

\section{Experiments and Results}
\label{sec:experiments-and-results}

We begin by introducing the datasets (Section \ref{subsec:datasets}), followed by detailing the featurization methods (Section \ref{subsec:featurization}), the experiments (Section \ref{subsec:experiments}), and finally reporting results (Section \ref{subsec:results}). 

\subsection{Datasets}
\label{subsec:datasets}
The statistics for the three datasets used in the experiments are summarized in Table \ref{tab:datasets}.
\begin{table}
\begin{center}
\scalebox{0.9}{
\begin{tabular}{lp{1.4cm}p{1.4cm}p{1.81cm}}
\toprule
Dataset & \# authors & mean words/auth & range words/auth \\
\midrule
\textsc{novel-9} & 9 &376,242 &124K-1M \\
\textsc{novel-50} & 50 &709,880 &184K-2.1M \\
\textsc{imdb62} & 62 &349,004 &9.8K-75K \\
\bottomrule
\end{tabular}}
\caption{Statistics for datasets.}
\label{tab:datasets}
\end{center}
\end{table}
\medskip

\noindent
\textbf{novel-9}.
This dataset was compiled by F\&H14: a collection of 19 novels by 9 nineteenth century British and American authors in the Project Gutenberg. To compare to F\&H14, we apply the same resampling method (F\&H14, Section 4.2) to correct the imbalance in authors by oversampling the texts of less-represented authors.
\medskip

\noindent
\textbf{novel-50}.
This dataset extends novel-9, compiling the works of 50 randomly selected authors of the same period. 
For each author, we randomly select 5 novels for a total 250 novels.
\medskip

\noindent
\textbf{IMDB62}.
IMDB62 consists of 62K movie reviews from 62 users (1,000 each) from the Internet Movie dataset, compiled by \newcite{Seroussi:11}. Unlike the novel datasets, the reviews are considerably shorter, with a mean of 349 words per text.

\subsection{Featurization}
\label{subsec:featurization}

As described in Section \ref{sec:background}, in both the GR and RST variants, from each input entry we start by obtaining an entity grid.
\medskip

\noindent
\textbf{CNN2-PV}.
We collect the probabilities of entity role transitions (in GR) or discourse relations (in RST) for the entries. Each entry corresponds to a probability distribution vector. 
\medskip

\noindent
\textbf{CNN2-DE}.
We employ two schema for creating discourse feature sequences from an entity grid. While we always read the grid by column (by a salient entity), we vary whether we track the entity across a  number of sentences (\textit{n} rows at a time) or across the entire document (one entire column at a time), denoted as \emph{local} and \emph{global} reading
respectively.

\begin{figure}[t]
\centering
\subfigure[local]{\includegraphics[width=27mm]{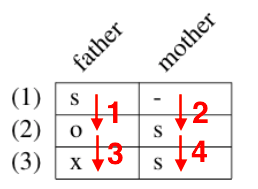}}
\subfigure[global]{\includegraphics[width=27mm]{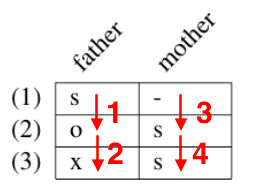}}
\vspace{-1.2em}
\caption{Two variants for creating sequences of grammatical relation transitions in an entity grid.}
\label{fig:globalLocal}
\vspace{-1.2em}
\end{figure}

For the GR discourse features, in the case of local reading, we process the entity roles one sentence pair at a time (Figure \ref{fig:globalLocal}, left).
For example, in processing the pair $(s_1,s_2)$, 
we find the first non-empty role $r_{11}$ for entity $E1$ in $s_1$. If $E1$ also has a non-empty role $r_{21}$ in the $s_2$, we collect the entity role transition $r_{11}r_{21}$. We then proceed to the following entity $E2$, until we process all the entities in the grid and move to the next sentence pair. 
For the global reading, we instead read the entity roles by traversing one column of the entire document at a time (Figure \ref{fig:globalLocal}, right). The entity roles in all the sentences are read for one entity: 
we collect transitions for all the non-empty roles (e.g., $\texttt{so}$, but not $\texttt{s-}$).

For the RST discourse features, we process non-empty discourse relations also through either local or global reading.
In the local reading, we read all the discourse relations in a sentence (a row) then move on to the next sentence.\footnote{We do not check the next sentences as in GR, because the discourse relations in one cell of the entity grid typically already capture relations beyond the sentence level.}
In the global reading, we read in discourse relations for one entity at a time. 
This results in sequences of discourse relations for the input entries. 

\subsection{Experiments}
\label{subsec:experiments}

\textbf{Baseline-dataset experiments}.
All the baseline-dataset experiments are evaluated on novel-9. As a comparison to previous work (F15), we evaluate our models using a pairwise classification task with GR discourse features. In her model, novels are partitioned into 1000-word chunks, and the model is evaluated with accuracy.\footnote{Averaged over all the author-author pair experiments.}
Surpassing F15's SVM model by a large margin, we then further evaluate the more difficult multi-class task, i.e., all-class prediction simultaneously, with both GR and RST discourse features and the more robust F1 evaluation.
In this multi-class task, we implement two SVMs to extend F15's SVM models:
(i) SVM2: a linear-kernel SVM which takes char-bigrams as input, as our CNNs, and (ii) SVM2-PV: an updated SVM2 which takes also probability vector features.

Further, we are interested in finding a performance threshold on the minimally-required input text length for discourse information to ``kick in''. 
To this end, we chunk the novels into different sizes: 200-2000 words, at 200-word intervals, and evaluate our CNNs in the multi-class condition. 
\medskip

\noindent
\textbf{Generalization-dataset experiments}.
To confirm that our models generalize, we pick the best models from the baseline-dataset experiments and evaluate on the novel-50 and IMDB62 datasets. 
For novel-50, the chunking size applied is 2000-word as per the baseline-dataset experiment results, and for IMDB62, texts are not chunked (i.e., we feed the models with the original reviews directly). 
For model comparison, we also run the SVMs (i.e., SVM2 and SVM2-PV) used in the baseline-dataset experiment.
All the experiments conducted here are multi-class classification with macro-averaged F1 evaluation.
\medskip

\begin{table}[t]
\begin{center}
\scalebox{0.9}{
\begin{tabular}{lc}
\toprule
\textsc{Model} & \textsc{Avg.Accuracy} \\
\midrule
Baseline & 49.8 \\
\midrule
SVM (LexSyn) & 85.5 \\
SVM (LexSyn-PV) & 86.4 \\
\midrule
CNN2 & 99.5 \\
CNN2-PV & \textbf{99.8} \\
\bottomrule
\end{tabular}}
\caption{Accuracy for pairwise author classification on the novel-9 dataset, using either a dumb baseline, an SVM with and without discourse to replicate F15, or a bigram-character CNN (CNN2) with and without discourse.}
\label{tab-num1:exp1}
\end{center}
\end{table}

\begin{table}[t]
\begin{center}
\scalebox{0.9}{
\begin{tabular}{llc}
\toprule
\textsc{Disc.Type} & \textsc{Model} & \textsc{F1} \\
\midrule
\multirow{2}{*}{None} & SVM2 & 84.9 \\
& CNN2 & 95.9 \\
\midrule
\multirow{4}{*}{GR} & SVM2-PV & 85.7 \\
& CNN2-PV & 96.1 \\
& CNN2-DE (local) & 97.0 \\
& CNN2-DE (global) & 96.9 \\
\midrule
\multirow{4}{*}{RST} & SVM2-PV & 85.9 \\
& CNN2-PV & 96.3 \\
& CNN2-DE (local) & 97.4 \\
& CNN2-DE (global) & \textbf{98.5} \\
\bottomrule
\end{tabular}}
\caption{Macro-averaged F1 score for multi-class author classification on the novel-9 dataset, using either no discourse (None), grammatical relations (GR), or RST relations (RST). These experiments additionally include the Discourse Embedding (DE) models for GR and RST.}
\label{tab-num2:exp1}
\end{center}
\end{table}

\noindent
\textbf{Model configurations}.
Following F15, we perform 5-fold cross-validation.
The embedding sizes are tuned on novel-9 (multi-class condition): 50 for char-bigrams; 20 for discourse features. 
The learning rate is 0.001 using the Adam Optimizer \citep{Kingma:14}.
For all models, we apply dropout regularization of 0.75 \citep{Srivastava:14}, and run 50 epochs (batch size 32).
The SVMs in the baseline-dataset experiments use default settings, following F15.
For the SVMs in the generalization-dataset experiments, we tuned the hyperparameters on novel-9 with a grid search, and found the optimal setting as: stopping condition \texttt{tol} is 1e-5, at a max-iteration of 1,500. 

\subsection{Results}
\label{subsec:results}
\begin{figure}[t]
\includegraphics[width=\linewidth]{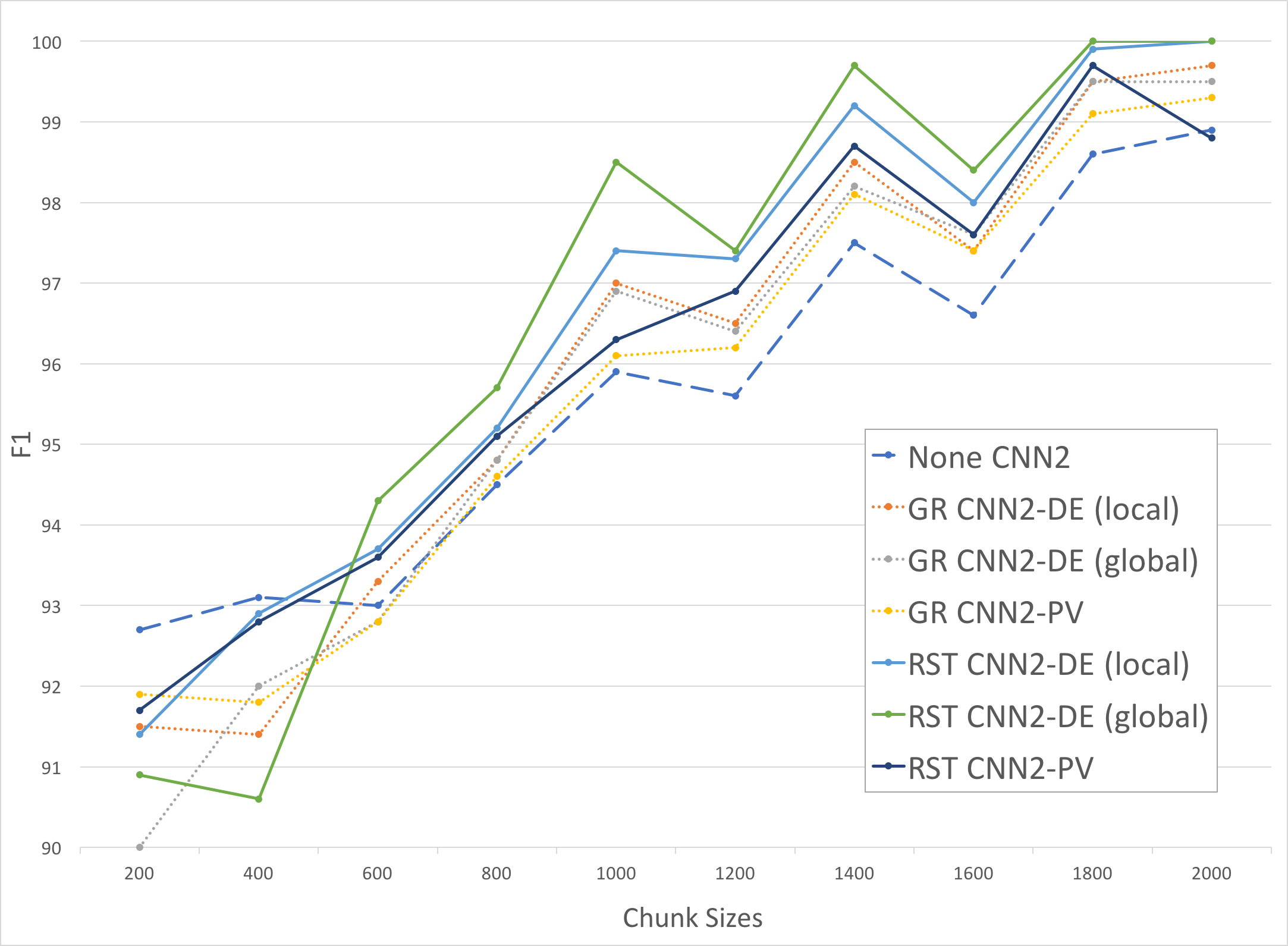}
\vspace{-1.6em}
\caption{Macro-averaged F1 score for multi-class author classification on the novel-9 dataset in varied chunk sizes.}
\label{fig:exp1-varying-sizes}
\end{figure}

\textbf{Baseline-dataset experiments}.
The results of the baseline-dataset experiments are reported in Table \ref{tab-num1:exp1}, \ref{tab-num2:exp1} and Figure \ref{fig:exp1-varying-sizes}. 
In Table \ref{tab-num1:exp1}, Baseline denotes the dumb baseline model which always predicts the more-represented author of the pair. 
Both SVMs are from F15, and we report her results.
SVM (LexSyn) takes character and word bi/trigrams and POS tags. 
SVM (LexSyn-PV) additionally includes probability vectors, similar to our CNN2-PV. 
In this part of the experiment, while the CNNs clear a large margin over SVMs, adding discourse in CNN2-PV brings only a small performance gain.

Table \ref{tab-num2:exp1} reports the results from the multi-class classification task, the more difficult task. 
Here, probability vector features (i.e., PV) again fail to contribute much.
The discourse embedding features, on the other hand, manage to increase the F1 score by a noticeable amount, with the maximal improvement seen in the CNN2-DE (global) model with RST features (by 2.6 points).
In contrast, the discourse-enhanced SVM2-PVs increase F1 by about 1 point, with overall much lower scores in comparison to the CNNs.
In general, RST features work better than GR features.

The results of the varying-sizes experiments are plotted in Figure \ref{fig:exp1-varying-sizes}.
Again, we observe the overall pattern that discourse features improve the F1 score, and RST features procure superior performance. 
Crucially, however, we note there is no performance boost below the chunk size of 1000 for GR features, and below 600 for RST features.
Where discourse features do help, the GR-based models achieve, on average, 1 extra point on F1, and the RST-based models around 2. 

\medskip

\noindent
\textbf{Generalization-dataset experiments}.
Table \ref{tab-num3:exp2} summarizes the results of the generalization-dataset experiments. 
On novel-50, most discourse-enhanced models improve the performance of the baseline non-discourse CNN2 to varying degrees. 
The clear pattern again emerges that RST features work better, with the best F1 score evidenced in the CNN2-DE (global) model (3.5 improvement in F1). 
On IMDB62, as expected with short text inputs (mean=349 words/review), the discourse features in general do not add further contribution. Even the best model CNN2-DE brings only marginal improvement, confirming our findings from varying the chunk size on novel-9, where discourse features did not help at this input size.
Equipped with discourse features, SVM2-PV performs slightly better than SVM2 on novel-50 (by 0.4 with GR, 0.9 with RST features).
On IMDB62, the same pattern persists for the SVMs:
discourse features do not make noticeable improvements (by 0.0 and 0.5 with GR and RST respectively). 

\begin{table}
\begin{center}
\scalebox{0.8}{
\begin{tabular}{llcc}
\toprule
\textsc{Disc. Type} & \textsc{Model} & \textsc{novel-50} & \textsc{IMDB62} \\
\midrule
\multirow{2}{*}{None} & SVM2 & 92.9 & 90.4 \\
& CNN2 & 95.3 & 91.5 \\
\midrule
\multirow{4}{*}{GR} & SVM2-PV & 93.3 & 90.4 \\
& CNN2-PV & 95.1 & 90.5 \\
& CNN2-DE (local) & 96.9 & 90.8 \\
& CNN2-DE (global) & 97.5 & 90.9 \\
\midrule
\multirow{4}{*}{RST} & SVM2-PV & 93.8 & 90.9 \\
& CNN2-PV & 95.5 & 90.7 \\
& CNN2-DE (local) & 97.7 & 91.4 \\
& CNN2-DE (global) & \textbf{98.8} & \textbf{92.0} \\

\bottomrule
\end{tabular}}
\caption{Macro-averaged F1 score for multi-class author classification on the large datasets, using either no discourse (None), grammatical relations (GR), or RST relations (RST).}
\label{tab-num3:exp2}
\vspace{-.5em}
\end{center}
\end{table}

\section{Analysis}
\label{sec:analysis}

\begin{table*}
\begin{center}
\scalebox{0.9}{
\begin{tabular}{cl}
\toprule
\textsc{Target embedding} & \textsc{Top neighbors} \\
\midrule
\texttt{explanation.N} & \texttt{interpretation.N, explanation.S, example.N,} \\ 
& \texttt{purpose.S, reason.N} \\
\texttt{background.N} & \texttt{circumstances.S, contrast.N, comparison.N,} \\ 
& \texttt{antithesis.S, elaboration.N} \\
\texttt{consequence.N} & \texttt{result.N, list.N, result.S,} \\
& \texttt{comment.N, summary.N} \\
\bottomrule
\end{tabular}}
\caption{Nearest neighbors of example embeddings with t-SNE clustering (top 5)}
\label{tab:tsne}
\end{center}
\end{table*}

\textbf{General analysis}.
Overall, we have  shown that discourse information can improve authorship attribution, but only when properly encoded.
This result is critical in demonstrating the particular value of discourse information, because typical stylometric features such as word $n$-grams and POS tags do {\it not} add additional performance improvements \citep{Ruder:16,Sari:17}.

In addition, the type of discourse information and the way in which it is featurized are tantamount to this performance improvement:
RST features provide overall stronger improvement, and the global reading scheme for discourse embedding works better than the local one.
The discourse embedding proves to be a superior featurization technique, as evidenced by the generally higher performance of CNN2-DE models over CNN2-PV models.
With an SVM, where the option is not available, we are only able to use relation probability vectors to obtain a very modest performance improvement. 

Further, we found an input-length threshold for the discourse features to help (Section \ref{subsec:results}).
Not surprisingly, discourse does not contribute on shorter texts.
Many of the feature grids are empty for these shorter texts-- either there are no coreference chains or they are not correctly resolved. 
Currently we only have empirical results on short novel chunks and movie reviews, but believe the finding would generalize to Twitter or blog posts.
\medskip

\noindent
\textbf{Discourse embeddings}.
It does not come as a surprise that discourse embedding-based models perform better than their relation probability-based peers.
The former (i) leverages the weight learning of the entire computational graph of the CNN rather than only the softmax layer, as the PV models do, and (ii) provides a more fine-grained featurization of the discourse information. Rather than merely taking a probability over grammatical relation transitions (in GR) or discourse relation types (in RST), in DE-based models we learn the dependency between grammatical relation transitions/discourse relations through the $w$-sized filter sweeps. 

To further study the information encoded in the discourse embeddings, we perform \texttt{t-SNE} clustering \cite{vanDerMaaten:08} on them, using the best performing model CNN2-DE (global). We examine the closest neighbors of each embedding, and observe that similar discourse relations tend to go together (e.g., \texttt{explanation} and \texttt{interpretation}; \texttt{consequence} and \texttt{result}). Some examples are given in Table \ref{tab:tsne}.
However, it is unclear how this pattern helps improve classification performance.
We intend to investigate this question in future work.
\medskip

\noindent
\textbf{Global vs. Local featurization}.
As described in Section \ref{subsec:featurization}, the global reading processes all the discourse features for one entity at a time, while the local approach reads one sentence (or one sentence pair) at a time.
In all the relevant experiments, global featurization showed a clear performance advantage (on average 1 point gain in F1). 
Recall that the creation of the grids (both GR and RST) depend on coreference chains of entities (Section \ref{sec:background}), and only the global reading scheme takes advantage of the coreference pattern whereas the local reading breaks the chains. 
To find out whether coreference pattern is the key to the performance difference, we further ran a probe experiment where we read RST discourse relations in the order in which EDUs are arranged in the RST tree (i.e., left-to-right), and evaluated this model on novel-50 and IMDB62 with the same hyperparameter setting. 
The F1 scores turned out to be very close to the CNN2-DE (local) model, at 97.5 and 90.9.
Based on this finding, we tentatively confirm the importance of the coreference pattern, and intend to further investigate how exactly it matters for the classification performance. 
\medskip

\noindent
\textbf{GR vs. RST}.
RST features in general effect higher performance gains than GR features (Table \ref{tab-num3:exp2}).
The RST parser produces a tree of discourse relations for the input text, thus introducing a ``global view.''
The GR features, on the other hand, are more restricted to a ``local view'' on entities between consecutive sentences. 
While a deeper empirical investigation is needed, one can intuitively imagine that identifying authorship by focusing on the local transitions between grammatical relations (as in GR) is more difficult than observing how the entire text is organized (as in RST).\footnote{Note that, however, it is simpler to extract GR features, as we rely solely on a high-performance dependency parser, which is widely available, whereas for RST features, we need gold RST-labeled training data, which incurs higher cost and potentially relatively limited generalizability.}

\section{Conclusion}
\label{sec:conclusion}
We have conducted an in-depth investigation of techniques that (i) featurize discourse information, and (ii) effectively integrate discourse features into the state-of-the-art character-bigram CNN classifier for AA. 
Beyond confirming the overall superiority of RST features over GR features in larger and more difficult datasets, we present a discourse embedding technique that is unavailable for previously proposed discourse-enhanced models.
The new technique enabled us to push the envelope of the current performance ceiling by a large margin. 

Admittedly, in using the RST features with entity-grids, we lose the valuable RST tree structure.
In future work, we intend to adopt more sophisticated methods such as RecNN, as per \newcite{Ji:17}, to retain more information from the RST trees while reducing the parameter size. Further, we aim to understand how discourse embeddings contribute to AA tasks, and find alternatives to coreference chains for shorter texts.




\bibliography{ijcnlp2017}
\bibliographystyle{ijcnlp2017}

\end{document}